\documentclass[10pt,twocolumn,letterpaper]{article}

\usepackage{cvpr}              

\usepackage{graphicx}
\usepackage{amsmath, amssymb, algorithm, url, natbib}
\usepackage{algpseudocode}
\usepackage{booktabs}
\usepackage{multirow}
\usepackage{capt-of} 
\definecolor{cvprblue}{rgb}{0.21,0.49,0.74}
\usepackage[pagebackref,breaklinks,colorlinks,allcolors=cvprblue]{hyperref}

\usepackage{arydshln}

\def\paperID{*****} 
\def\confName{CVPR}
\def\confYear{2026}

\title{A Contrastive Learning Framework Empowered by Attention-based Feature Adaptation for Street-View Image Classification}

\author{Qi You$^{*}$ \\
SpaceTimeLab\\
University College London\\
\and
Yitai Cheng$^{*}$ \\
SpaceTimeLab\\
University College London\\
\and
Zichao Zeng\\
3DIMPact \& SpaceTimeLab\\
University College London\\
\and
James Haworth$^\dagger$\\
SpaceTimeLab\\
University College London\\
}

\begin{document}
\maketitle
\renewcommand{\thefootnote}{}
\footnotetext{
$^{*}$ Equal contribution.
}
\footnotetext{
$^\dagger$ Corresponding author: j.haworth@ucl.ac.uk}
\renewcommand{\thefootnote}{\arabic{footnote}}

\vspace{0.75em} 

\begin{abstract}
    Street-view image attribute classification is a vital downstream task of image classification, enabling applications such as autonomous driving, urban analytics, and high-definition map construction. It remains computationally demanding whether training from scratch, initialising from pre-trained weights, or fine-tuning large models. Although pre-trained vision-language models such as CLIP offer rich image representations, existing adaptation or fine-tuning methods often rely on their global image embeddings, limiting their ability to capture fine-grained, localised attributes essential in complex, cluttered street scenes. 
    To address this, we propose CLIP-MHAdapter, a variant of the current lightweight CLIP adaptation paradigm that appends a bottleneck MLP equipped with multi-head self-attention operating on patch tokens to model inter-patch dependencies. With approximately 1.4 million trainable parameters, CLIP-MHAdapter achieves superior or competitive accuracy across eight attribute classification tasks on the Global StreetScapes dataset, attaining new state-of-the-art results while maintaining low computational cost.
    The code is available at \href{https://github.com/SpaceTimeLab/CLIP-MHAdapter}{https://github.com/SpaceTimeLab/CLIP-MHAdapter}.
\end{abstract}    
\section{Introduction}
\label{sec:intro}
Street-view imagery (SVI) has rapidly emerged as a transformative geospatial data source, reshaping how we sense, measure, and understand the complexity of urban environments. Unlike remote sensing, which captures images from a vertical, top-down perspective, SVI provides a horizontal, human-centred viewpoint that reflects how streets are experienced at ground level. This unique perspective is increasingly leveraged for applications such as high-definition (HD) map construction, urban planning, and environmental monitoring. Commercial SVI services, including Google Street View, Baidu Maps, and Tencent Street View, provide standardised, high-quality panoramas collected by vehicle-mounted cameras with supporting sensors like LiDAR, ensuring consistent metadata and uniform coverage. However, these sources tend to have restrictions on their use \cite{helbich_use_2024}. Crowd-sourced platforms such as Mapillary and KartaView provide open alternatives for SVI, but have a greater variation in quality, format and coverage due to lack of standardisation \cite{HOU2022103094}. A key challenge in utilising these heterogeneous SVI sources lies in their lack of reliable metadata to filter images by contextual attributes, such as weather, lighting, or capture platform, making large-scale analysis and multi-city comparisons prone to bias and noise. To unlock the full potential of open source SVI for robust urban analytics, systematic attribute classification is essential, enabling image filtering, dataset curation, and reliable downstream tasks.

Street-view image attribute classification can be approached using conventional image classification pipelines or by adapting modern vision-language models. Training deep networks from scratch or fine-tuning large pre-trained vision language models can be computationally intensive, especially when scaling to millions of SVI. Recent progress in vision-language models, such as Contrastive Learning-based Image Pretraining (CLIP) \cite{radford2021learning}, provides rich visual representations that generalise across domains and can be leveraged for downstream SVI tasks. This can be achieved by Parameter-Efficient Adaptation (PEA). However, most existing PEA strategies for CLIP, such as \cite{zhouConditionalPromptLearning2022} and \cite{gao2024clip}, operate primarily on global image embeddings. While effective for coarse-grained scene recognition, this approach is inherently limited for fine-grained attribute classification in complex urban scenes, where key cues may be spatially localised or partially occluded. For example, detecting a reflection on a car window or a weather-related condition like fog often requires reasoning over local patch-level features and their spatial relationships, which global embeddings may fail to capture. Consequently, achieving both high accuracy and computational efficiency for SVI attribute classification remains an open challenge.

To address these limitations, we propose CLIP-MHAdapter, a novel variant of the CLIP adaptation paradigm that employs a lightweight and scalable adaptation strategy specifically tailored for SVI attribute classification. 
The method freezes the CLIP backbone and appends a bottleneck multi-layer perceptron (MLP) equipped with multi-head self-attention operating on patch tokens, enabling the model to capture inter-patch dependencies and fine-grained spatial cues without the high cost of full model fine-tuning.
Extensive experiments on the Global StreetScapes (GSS) dataset \cite{hou2024global}, which covers eight attribute classification tasks, show that CLIP-MHAdapter achieves competitive or superior performance compared to the fully trained baseline released with the dataset, while reducing training cost and facilitating deployment on resource-constrained edge devices.


To summarise, our contributions are listed as follows:
\begin{itemize}
\item We propose \textbf{CLIP-MHAdapter}, a variant of the CLIP adaptation paradigm that integrates a bottleneck MLP with multi-head self-attention to effectively capture inter-patch dependencies and fine-grained spatial cues in street-view imagery.
\item Our approach achieves a better efficiency–accuracy trade-off, delivering higher accuracy than existing CLIP adaptation methods with only a moderate increase in trainable parameters, and remaining much smaller than full fine-tuning.
\item We conduct extensive experiments on the Global StreetScapes (GSS) dataset, covering eight attribute classification tasks, where CLIP-MHAdapter achieves competitive or superior accuracy to the fully trained baseline while significantly reducing training cost and facilitating deployment on resource-constrained devices.
\end{itemize}

By enabling efficient and accurate SVI attribute classification, our method paves the way for scalable, reliable, and fine-grained urban analytics.


\begin{figure*}[t]
  \centering
  \includegraphics[width=0.95\textwidth]{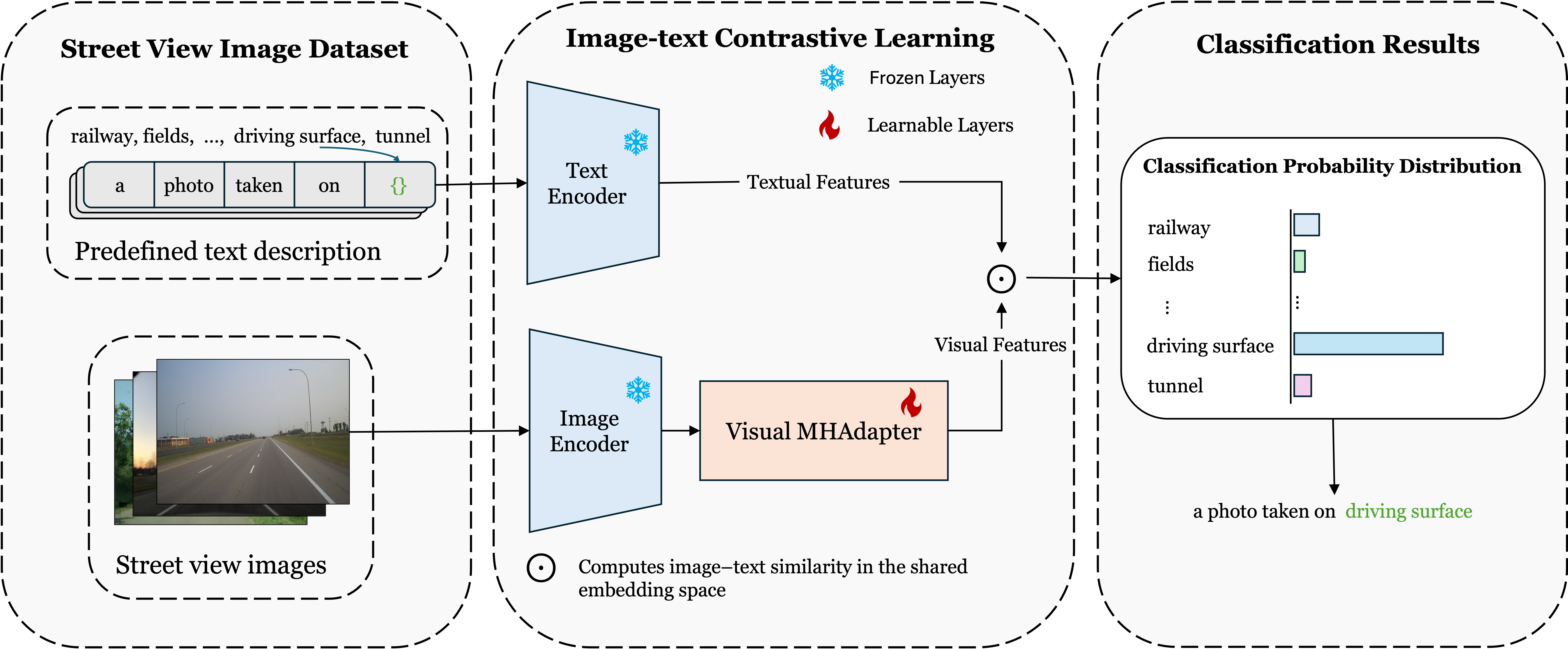}
  \caption{The CLIP-MHAdapter framework. A visual MHAdapter module is integrated downstream of the pretrained image encoder, enabling task-specific adaptation while preserving the representational strengths of the pretrained CLIP backbone \cite{radford2021learning}. }
  \label{fig:teaser}
\end{figure*}
\section{Related Work}
\label{sec:related work}
\paragraph{Street-view Image Analysis} SVI has become a prominent data source for urban analytics. A review of SVI research by \cite{biljecki2021street} found over 600 uses of SVI in academic literature, with a diverse range of application areas including spatial data infrastructure, health and wellbeing, urban perception and transport and mobility. Automated analysis of SVI at scale typically requires computer vision (CV) techniques such as image classification, object detection and semantic segmentation and the majority of studies to date rely on deep learning models such as convolutional neural networks (CNNs) to carry out these tasks. A review by \cite{he_urban_2021} found deep learning to be a common method in tasks ranging from thermal environment analysis to neighbourhood environment perception. Typical architectures include visual geometry group (VGG) variants, ResNet, SegNet, DeepLab V3+ and, more recently, vision transformers \cite{dai_street_2024, zhang_urban_2024}.

While traditional deep learning exhibits strong performance across a wide range of CV tasks, it typically requires large labelled datasets for training, limiting its applicability to heterogeneous SVI tasks.  Furthermore, pre-trained deep learning models tend to be trained on benchmark datasets collected in certain locales, such as the United States, limiting transferability to different contexts. The majority of academic research uses Google Street View (GSV), which has highly standardized data. However, application of CV to GSV is technically prohibited according to Google's terms of use, which poses a risk to SVI pipelines that rely on GSV \cite{helbich_use_2024}. An alternative to proprietary sources is open-source SVI, such as Meta Mapillary and KartaView. Due to their crowd-sourced nature, they are of varying quality, having been collected by a diverse user group using a variety of devices and platforms (e.g. vehicle, bicycle, pedestrian). Typically, a researcher will need to filter high quality images from the larger set before analysing open-source SVI. Often, the metadata may not contain the necessary information to accomplish this task, meaning CV pipelines are required, as demonstrated by \cite{hou2024global}. In this case, training a CV model to classify image quality-related attributes would require a considerable volume of labelled data under traditional deep learning paradigms, as these methods are inherently data-intensive and rely on large annotated datasets to generalize effectively.

\paragraph{Vision Language Models} Recently, vision language models (VLMs) have emerged as a new paradigm for CV tasks, which can be generally described as multimodal systems that jointly learn representations from both visual and textual inputs. Visual-language joint learning models introduced by \cite{tanLXMERTLearningCrossModality2019, luViLBERTPretrainingTaskAgnostic2019} require multi-stage pretraining with different designated tasks.
In contrast with previous work, \cite{desaiVirTexLearningVisual2021} explored pretraining a CNN via only image captioning task and demonstrated a competitive performance on downstream tasks with those obtained via supervised classification and self-supervised learning on ImageNet \cite{dengImageNetLargescaleHierarchical2009}, one of the generic benchmark datasets for various vision tasks. More recent VLMs \cite{radford2021learning, jiaScalingVisualVisionLanguage2021, yaoFILIPFINEGRAINEDINTERACTIVE2022, zhaiLiTZeroShotTransfer2022} are often trained with contrastive learning approach \cite{chenSimpleFrameworkContrastive2020}, large-scale multi-modal data \cite{schuhmannLAION5BOpenLargescale2022, xiaoFlorence2AdvancingUnified2024}, and more powerful encoders like Transformers \cite{vaswani2017attention}. CLIP is one of the representative models \cite{radford2021learning}, which jointly trains two neural encoders for images and text with a contrastive learning objective to align paired visual–language representations. Leveraging 400 million image–text pairs for pretraining, CLIP exhibits strong zero-shot image recognition performance.


\paragraph{Adaptation Strategies for CLIP}
Large-scale vision foundation models are typically pre-trained on generic datasets which capture diverse object- and scene-level semantics but differ significantly from the complex, cluttered image data encountered in real-world applications like SVI.
Several lightweight adaptation strategies have emerged to exploit pre-trained CLIP for downstream tasks without full model fine-tuning. Linear probing trains a simple classifier on frozen features, offering low computation but limited expressivity. In \cite{radford2021learning}, researchers validated the effectiveness of linear probing by freezing the pretrained CLIP vision encoder and training only a linear classifier on its extracted features. 
Rather than adding extra fine-tuning modules, prompt learning approaches, such as \cite{zhouLearningPromptVisionLanguage2022, zhouConditionalPromptLearning2022,  yaoVisualLanguagePromptTuning2023, bulatLASPTexttoTextOptimization2023}, parameterize textual prompts as learnable vectors to enhance vision–language alignment. \cite{khattakMaPLeMultimodalPrompt2023} extended the idea of prompt learning for VLMs by adding extra learnable context embeddings to both image and text branch of CLIP.
In contrast, adapter-based approach such as \cite{gao2024clip, yangMMAMultiModalAdapter2024}, pioneers efficient and effective model adaptation by injecting lightweight trainable modules into the frozen CLIP backbone. Compared to linear-probe and prompt learning-based methods, CLIP-Adapter \cite{gao2024clip} was proved to achieve better performance under various few-shot learning settings on 11 image classification datasets, while \cite{yangMMAMultiModalAdapter2024} can achieve better or competitive performance in the base-to-novel, Cross-Dataset, and domain generalization settings with 16-shot training. Despite the outstanding performance on multiple general datasets, these adaptation training strategies primarily focus on global representations encoded by CLIP and tend to overlook localized image features, which limits their ability to capture fine-grained visual details essential for detecting subtle SVI attributes, such as reflections, shadows, and other localized effects. Consequently, these approaches remain under-explored in the context of street-view imagery.

\section{Methodology}
\label{sec:methodology}
In this section, we present CLIP-MHAdapter, a lightweight vision-language fine-tuning framework for downstream image classification tasks requiring fine-grained visual understanding. Our method builds on CLIP by introducing a multi-head, self-attention-enhanced visual adapter that refines image features, while keeping the text encoder entirely frozen to maintain efficiency. This design is not limited to street-view imagery (SVI) but naturally benefits scenarios where subtle local cues and global context must be jointly leveraged.
We first provide an overview of the image encoder of CLIP-MHAdapter in Section~\ref{subsec:clip}, then introduce our residual multi-head self-attention adapter (MHAdapter) in Section~\ref{subsec:adapter}, followed by the text encoder in Section~\ref{subsec:text}. Finally, we describe the imbalance-aware weighting strategy for model training in Section~\ref{subsec:weighting}.

\subsection{Global-Local Image Encoder} \label{subsec:clip}
Image classification pipelines with deep neural networks typically consist of two stages: feature extraction and classification. Let $\mathbf{I} \in \mathbb{R}^{H \times W \times 3}$ denote an image with height $H$, width $W$, and three RGB channels. Following the Vision Transformer (ViT) paradigm, the image is first divided into non-overlapping patches~\cite{dosovitskiy2020image}. Each patch is then linearly projected into an embedding and subsequently encoded by the backbone as a token sequence, with a class token prepended to serve as a global representation of the entire image~\cite{dosovitskiy2020image}. Formally, the ViT backbone outputs
\begin{equation}
\label{eq:vit}
    \mathbf{f} = \text{ViT}(\mathbf{I}),
\end{equation}
where $\mathbf{f} \in \mathbb{R}^{(N+1) \times D}$, $D$ is the feature dimensionality and $N$ is the number of image patches. The first token $\mathbf{f}_{0} \in \mathbb{R}^D$ serves as the global image feature, while the remaining $N$ tokens $\mathbf{f}_{1:N} \in \mathbb{R}^{N \times D}$ correspond to the patch-level features that preserve local spatial details. Unlike CLIP-Adapter which only uses the global feature (i.e., $\mathbf{f}_{0}$) to finetune their adaptation layers, our CLIP-MHAdapter exploits the local spatial features (i.e., $\mathbf{f}_{1:N}$) while maintaining the awareness of the global features, achieved by a multi-head feature adaptation module whose detail will be introduced in \ref{subsec:adapter}. The adapted visual feature is denoted as $\mathbf{f}^{*}$.

A linear classifier subsequently projects $\mathbf{f}^{*}$ into a $K$-dimensional logit space using a weight matrix $\mathbf{W} \in \mathbb{R}^{D\times K}$, followed by a temperature-scaled Softmax function to obtain class probabilities:
\begin{equation}\label{eq:prob}
p_i = \frac{\exp \big( \mathbf{W}_i^\top \mathbf{f}^{*} / \tau \big)}
           {\sum_{j=1}^{K} \exp \big( \mathbf{W}_j^\top \mathbf{f}^{*}/ \tau \big)}
,\end{equation}
where $\tau > 0$ is the Softmax temperature, $\mathbf{W}_i$ is the classifier weight for class $i$, and $p_i$ the probability for class $i$ satisfies $\sum_{i=1}^K p_i = 1$.
\subsection{Multi-Head Feature Adaptation} \label{subsec:adapter}
\begin{figure*}[t]
  \centering
  \includegraphics[width=0.95\textwidth]{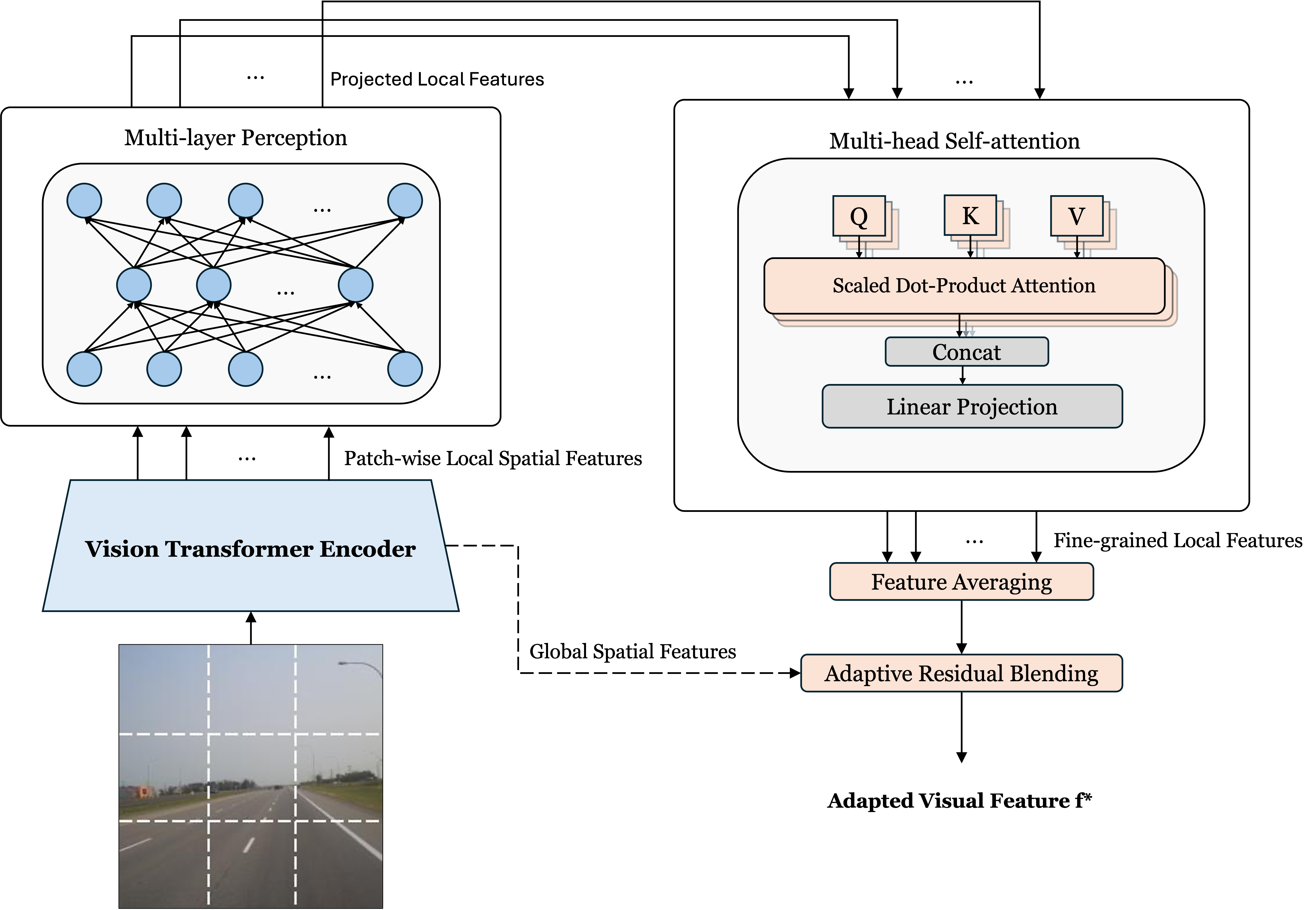}
  \caption{Details of the Multi-Head Feature Adaptation Module. Note that the input images are partitioned into standard 16 × 16 patches before being fed into the Vision Transformer encoder \cite{dosovitskiy2020image}. A larger patch size is shown here solely for clarity of illustration.}
  \label{fig:detailed-adapter}
\end{figure*}

To efficiently adapt for downstream image classification tasks that demand fine-grained visual understanding, particularly within the domain of street-view imagery (SVI), we append a \textbf{m}ulti-\textbf{h}ead self-attention enhanced feature \textbf{a}daptation module (MHA($\cdot$)) to the visual encoder while keeping both visual and textual encoder backbones frozen during fine-tuning. By operating on the extracted visual representations, this adapter enhances fine-grained discriminative capability for SVI image classification with minimal additional parameters, providing a more specialised view of SVI without modifying the backbone.

Concretely, given an image $\mathbf{I}$ and a category set $\{C_i\}_{i=1}^K$, the image feature $\mathbf{f}$ and classifier weight $\mathbf{W}$ from the frozen CLIP backbone is calculated following Equation~\ref{eq:vit} and Equation~\ref{eq:text-encoder-clip}. Patch-level embeddings $\mathbf{f}_{1:N} \in \mathbb{R}^{N\times D}$, extracted from frozen visual encoder, are processed by a lightweight bottleneck Multi-Layer Perceptron (MLP) projection to encourage discriminative adaptation without heavy computation:
\begin{equation}\label{eq:mlp}
\mathbf{X}_\text{av} =  \mathrm{ReLU}\big(\mathbf{f}_{1:N}\mathbf{W}_v^1\big)\mathbf{W}_v^2 
\end{equation}
where $\mathbf{W}_v^1 \in \mathbb{R}^{D \times d_b}$ and $\mathbf{W}_v^2 \in \mathbb{R}^{d_b \times D}$ are learnable weights from the bottleneck projection, $d_b \ll D$ is the bottleneck dimension, and the output $\mathbf{X}_\text{av} \in \mathbb{R}^{N \times D}$ comprises $N$ refined token features $\{\mathbf{x}_{n}\}^N_{n=1}$, each $\mathbf{x}_{n}$ being the projection of the original CLIP embedding for the $n$-th image patch. To stabilize these features, we apply layer normalisation \citep{baLayerNormalization2016} along the feature dimension independently for each token. For the $n$-th token, layer normalisation computes:
\begin{equation}
\begin{aligned}
\mu_n &= \frac{1}{D} \sum_{j=1}^{D} \mathbf{x}_{n,j}, \\
\sigma^2_n &= \frac{1}{D} \sum_{j=1}^{D} (\mathbf{x}_{n,j} - \mu_n)^2, \\
\mathbf{z}_n &= \boldsymbol{\gamma} \cdot 
\frac{\mathbf{x}_n - \mu_n}{\sqrt{\sigma_n^2 + \epsilon}} + \boldsymbol{\beta}
\end{aligned}
\end{equation}
where $\gamma, \beta \in \mathbb{R}^{D}$ are learnable per-dimension affine parameters shared across all tokens, and $\epsilon > 0$ the small constant ensures numerical stability. By stacking all $N$ token outputs, the layer-normalised result $\mathbf{X}_{\text{ln}} = [\,\mathbf{z}_1, \dots, \mathbf{z}_N\,] \in \mathbb{R}^{N \times D}.$
Subsequently, a multi-head self-attention (MHSA) mechanism is applied to capture inter-token dependencies and spatially localised relationships. 
Here, we adopt the formulation where the query ($\mathbf{Q}$), key ($\mathbf{K}$), and value ($\mathbf{V}$) matrices are all derived from the layer-normalised token sequence $\mathbf{X}_\text{ln}$ to form self-attention:
\begin{align}
\mathbf{Q} &= \mathbf{X}_\text{ln} \mathbf{W}_Q, &
\mathbf{K} &= \mathbf{X}_\text{ln} \mathbf{W}_K, &
\mathbf{V} &= \mathbf{X}_\text{ln} \mathbf{W}_V
\end{align}
where $\mathbf{W}_Q, \mathbf{W}_K, \mathbf{W}_V \in \mathbb{R}^{D \times D}$ are learnable projection matrices. In multi-head attention, the input dimension $D$ is split across $H$ heads with each head processing a subspace of dimension $d_k = D/H$ to allow parallel computation while maintaining total parameter efficiency. The attention output for the $h$-th head is computed as:
\begin{equation}\label{eq:mhsa-head}
\mathrm{head}^{(h)} 
= \mathrm{Softmax}\Bigg(
\frac{\mathbf{Q}^{(h)} {\mathbf{K}^{(h)}}^\top}{\sqrt{d_k}}
\Bigg) \mathbf{V}^{(h)} 
\end{equation}
Where $\mathrm{head}^{(h)}\: \in \mathbb{R}^{N \times d_k}$. All $H$ heads operate in parallel and their outputs are concatenated and linearly projected by a learnable projection layer:
\begin{equation}\label{eq:mhsa}
\mathbf{X}_\text{MHSA} 
:=  \operatorname{Concat}\big(\mathrm{head}^{(1)}, \ldots, \mathrm{head}^{(H)}\big) \mathbf{W}_O
\end{equation}
where $\mathbf{W}_O \in \mathbb{R}^{D \times D}$ is the output projection matrix and $\mathbf{X}_\text{MHSA}\: \in \mathbb{R}^{N \times D}$ is the MHSA ouput. A dropout operation with probability $p$ is applied to the multi-head self-attention output to mitigate overfitting, introducing stochastic regularisation:
\begin{equation}
\begin{aligned}
&\mathbf{M} \sim \mathrm{Bernoulli}(1-p), 
\\
&\mathbf{X}_\text{drop} =  \frac{\mathbf{M} \odot \mathbf{X}_\text{MHSA}}{1-p}
\end{aligned}
\end{equation}
where $\mathbf{M}$ is a Bernoulli mask of the same shape as $\mathbf{X}_\text{MHSA}$, and the scaling vector $1 / (1 - p)$ preserves the expected magnitude of the activations during training. To obtain a unified image-level embedding, the patch-level tokens are aggregated by feature averaging:
\begin{equation}
\mathbf{x}_\text{mean} = \frac{1}{N} \sum_{i=1}^{N} \mathbf{X}_{i, \text{drop}}
\end{equation}
where $\mathbf{x}_\text{mean}\in \mathbb{R}^{D}$ is effectively the output of the multi-head self-attention adaptation module $\mathrm{MHA}(\mathbf{f}_{1:N})$.

To preserve CLIP's original generalisation capabilities during downstream adaptation, the adapted features are blended with frozen CLIP embeddings via a residual blending mechanism:
\begin{equation}\label{eq:blend}
    \mathbf{f}^* = \alpha \times \mathrm{MHA}(\mathbf{f}_{1:N}) + (1 - \alpha) \times \mathbf{f}_{0}, 
\end{equation}
where $\alpha \in [0,1]$ controls the contribution of the adapted feature. 
The new obtained image feature $\mathbf{f}^*$ and classifier weight $\mathbf{W}$ are used to compute probability $p_i$ for class $i$ according to equation \ref{eq:prob}, and the image category is predicted by selecting class $\hat{i} = \mathop{\mathrm{arg\,max}}_{i} p_i$ that has the highest probability.

\subsection{Text Encoder}
\label{subsec:text}
In line with prior work \citep{radford2021learning, gao2024clip, zhouLearningPromptVisionLanguage2022}, CLIP-MHAdapter adopts a visual–language contrastive representation learning framework. As in the general classification pipeline summarised in Section~\ref{subsec:clip}, CLIP-MHAdapter generates classifier weights from text prompts rather than learning them from labelled images. Given a category set $\{C_i\}_{i=1}^K$ of natural language class names and a pre-defined prompt template $H$, CLIP-MHAdapter  constructs a textual prompt by concatenating $H$ with each class name $C_i$, then tokenizes it, and feeds it into its language encoder to produce the corresponding classifier weight:
\begin{equation}\label{eq:text-encoder-clip}
\mathbf{W}_i = \text{Text-Encoder}\Big(\text{Tokenizer}\big([H; C_i]\big)\Big)
\end{equation}
where $[H; C_i]$ denotes the concatenation of the prompt template and the class name.
Thus, the generated text-based classifier weights $\mathbf{W}_i$ enable the computation of class prediction probabilities $p_i$ using equation \ref{eq:prob}.

\subsection{Imbalance-Aware Weighting}
\label{subsec:weighting}

The cross-entropy loss is used for our model training and is defined as below:
\begin{equation}
    \mathcal{L}_\theta = -\frac{1}{N_b} \sum_{i=1}^{N_b}
    \log \frac{\exp\left( W_i^\top f_i / \tau \right)}
    {\sum_{j=1}^{K} \exp\left( W_j^\top f_i / \tau \right)},
\end{equation}
where $K$ is the number of classes and, $N_b$ is the number of training examples in a batch, and $\theta $ represents all learnable parameters of CLIP-Adapter.

To address class imbalance, we incorporated an inverse-frequency weighting strategy during training. This approach adjusts each class’s contribution to the loss function without altering the data distribution, thereby mitigating evaluation bias.

\paragraph{Inverse-Frequency Weighting:}
Assigns weights inversely proportional to class frequencies:
\begin{equation}
w_{\text{inv}}(c) = \frac{1/n_c}{\sum_{j=1}^{K} 1/n_j},
\end{equation}
where $n_c$ is the number of samples in class $c$. 

This imbalance-aware weighting mitigates class bias while preserving data integrity, leading to the training loss for our proposed MHAdapter:

\begin{equation}
\mathcal{L}_{\theta}
= -
\frac{
    \displaystyle
    \sum_{i=1}^{N_b}
    w_{\text{inv}}(c_i)\,
    \log
    \frac{
        \exp\!\left( W_i^\top f_i / \tau \right)
    }{
        \sum_{j=1}^{K} \exp\!\left( W_j^\top f_i / \tau \right)
    }
}{
    \displaystyle
    \sum_{i=1}^{N_b} w_{\text{inv}}(c_i)
}.
\end{equation}

\section{Data}
\label{sec:data}
\begin{table*}[ht]
\centering
\caption{
Overview of the manually labelled subset of GSS dataset \cite{hou2024global}. }
\begin{tabular}{l c r r l}
\toprule
\textbf{Attribute} & \textbf{No. of classes} &
\multicolumn{2}{c}{\textbf{Number of images}} &
\textbf{Possible categories} \\
\cmidrule(lr){3-4}
& & \textbf{train} & \textbf{test} & \\
\midrule
Platform            & 6 & 25,407  & 7,311 & driving/walking/cycling surface, railway, fields, tunnel \\
Weather             & 5 & 27,771  & 8,068 & clear, cloudy, rainy, snowy, foggy \\
View direction      & 2 & 7,632  & 2,012 & front/back, side \\
Lighting condition  & 3 & 9,380  & 3,079 & day, night, dusk/dawn \\
Panoramic status    & 2 & 8,372  & 2,172 & true, false \\
Quality             & 3 & 8,199  & 2,107 & good, slightly poor, very poor \\
Glare               & 2 & 8,089 & 2,115 & true, false \\
Reflection          & 2 & 8,112 & 2,119 & true, false \\
\bottomrule
\end{tabular}
\label{tab:data}
\end{table*}

Crowdsourced datasets refer to collections of data generated and contributed voluntarily by individuals using heterogeneous devices, rather than being captured through controlled or standardised commercial acquisition pipelines.
Such datasets offer substantial benefits, including broader geographic coverage, greater diversity of viewpoints and environmental conditions, improved temporal granularity, and more permissive licensing that makes them highly suitable for research.
Therefore, in this study, we adopt the Global StreetScapes (GSS) dataset \cite{hou2024global}, a large-scale and richly annotated crowdsourced street-view imagery resource that enables fine-grained urban visual analysis.

\subsection{Global StreetScapes (GSS)}
The GSS dataset is a comprehensive, openly accessible street-view imagery corpus comprising ten million images collected from 688 cities across 210 countries and territories. Each image is enriched with over 300 contextual, geographical, temporal, semantic and perceptual attributes, providing an extensive foundation for urban analytics and street-view understanding.

The imagery and associated metadata originate from two major crowdsourced street-view platforms, Mapillary and KartaView, from which raw images and native metadata (e.g., geographic coordinates, timestamps, projection types and sequence identifiers) were obtained via their publicly available APIs. These primary attributes were then further augmented using external geospatial sources such as OpenStreetMap, GADM, GHSL and the H3 spatial indexing system, as well as computer-vision models for semantic, contextual and perceptual inference. Moreover, 10,000 street-view images were randomly sampled approximately and manually labelled to provide reliable ground-truth annotations across the eight contextual attributes: panoramic status, lighting condition, view direction, weather, platform, image quality, and the presence of glare and reflections \cite{hou2024global} . To further mitigate class imbalance in the lighting, platform and weather attributes, more than 50,000 additional images manually curated from the Mapillary and KartaView web applications were incorporated into the labelled subset.

In this work, we focus specifically on aforementioned labelled subset with eight contextual attributes, selected for their relevance in assessing the suitability of an image for SVI classification tasks, as summarised in Table~\ref{tab:data}. 

\subsection{Data Pre-Processing}

The dataset was partitioned into training and testing subsets and subsequently used to train and evaluate the proposed CLIP-MHAdapter model alongside comparison baselines. All images were pre-processed in accordance with the conventions established for CLIP \citep{radford2021learning} to ensure comparability with existing methods. Each image was resized to 224×224 pixels using bicubic interpolation, followed by channel normalisation and standard data-augmentation techniques, including random cropping and random horizontal flipping, to enhance scale, positional and viewpoint invariance. A stratified sampling procedure was employed to hold out 20\% of the training data as a validation set, preserving class distributions across subsets. This validation set was used to enable early stopping, thereby reducing overfitting and improving generalisation performance \citep{prechelt1998automatic}.

\section{Experimental Setup}
\label{sec:experiments}
To rigorously evaluate the effectiveness of CLIP-MHAdapter for contextual attribute recognition in crowdsourced street-view imagery, we conduct a comprehensive suite of experiments on the Global StreetScapes dataset. Our experimental design examines four key dimensions: (i) comparison against a diverse set of baselines encompassing zero-shot transfer, parameter-efficient adaptation, and high-capacity vision transformers; (ii) controlled implementation settings to ensure reproducibility and fair comparison; (iii) systematic evaluation using metrics that account for both overall accuracy and class imbalance; and (iv) analysis of performance across heterogeneous visual conditions inherent to real-world imagery. Together, these components provide a robust empirical foundation for assessing the advantages of our proposed method.

\subsection{Comparison Models}\label{sec:baselines}
We benchmark it against a diverse collection of baselines spanning zero-shot vision–language transfer, parameter-efficient CLIP adaptations, and contemporary vision transformers. Together, these baselines represent a broad spectrum of adaptation strategies, offering complementary perspectives on urban street-view classification. The comparison models are summarised as follows:
\paragraph{Zero-Shot Transfer}
\begin{itemize}
    \item \textbf{Zero-Rule Classifier (ZeroR):} A trivial baseline that always predicts the majority class. This model requires no training and provides a minimal reference point, establishing the lower bound of classification performance.
    \item \textbf{Zero-Shot CLIP} \cite{radford2021learning}: The original CLIP model applied directly without finetuning. By mapping image and text features into a joint embedding space, Zero-Shot CLIP is a strong baseline for transfer learning, demonstrating how far pretrained vision-language alignment alone can generalize without domain-specific. 
\end{itemize}

\paragraph{Parameter-Efficient Adaptation}
\begin{itemize}
    \item \textbf{CLIP + Linear Probe} \cite{radford2021learning}: A conventional adaptation strategy where a lightweight linear classifier is trained on frozen CLIP embeddings. This approach isolates the benefit of simple supervised alignment while preserving pretrained representations, serving as a mid-level benchmark between zero-shot and more sophisticated finetuning methods.
    \item \textbf{CoOp} \cite{zhouLearningPromptVisionLanguage2022}: A prompt-learning method that optimizes soft textual context vectors. Instead of hand-crafted prompts, CoOp learns a sequence of continuous embeddings, concatenated with the tokenised class label and passed through CLIP’s text encoder. This strategy adapts CLIP to downstream tasks by tuning only prompt tokens while keeping the backbone frozen, balancing efficiency and accuracy. 
    \item \textbf{CLIP-Adapter} \cite{gao2024clip}: A lightweight adaptation framework that introduces residual adapter layers into the image and text branches. The adapter blends pretrained features with task-specific transformations via a learnable gating mechanism. This design is parameter-efficient, stable, and enables CLIP to specialize to downstream distributions. 
\end{itemize}

\paragraph{Vision Transformer}
\begin{itemize}
    \item \textbf{MaxViT} \cite{tu2022maxvit}: A vision transformer architecture employing multi-axis attention and block-wise factorisation. Unlike CLIP-based methods, MaxViT is a pure vision backbone trained from scratch or with ImageNet pretraining. This model provides a comparison against a high-capacity transformer trained for visual recognition rather than cross-modal alignment. 
\end{itemize}

\subsection{Implementation Details}\label{sec:implementation}
\begin{table*}[ht]
\centering
\caption{Optimal hyperparameters combinations for CLIP-MHAdapter per Attribute.}
\begin{tabular}{lccc}
\toprule
\textbf{Attribute} & \textbf{Class Weights} & \textbf{Blend Ratio} & \textbf{No. of Attention Heads} \\
\midrule
Platform            & Uniform & 0.8 & 4  \\
Weather             & Uniform & 0.2 & 8  \\
View direction      & Inverse & 0.8 & 4  \\
Lighting condition  & Inverse & 0.8 & 8  \\
Panoramic status    & Uniform & 0.2 & 4  \\
Quality             & Inverse & 0.2 & 16 \\
Glare               & Uniform & 0.8 & 4  \\
Reflection          & Inverse & 0.8 & 8  \\
\bottomrule
\end{tabular}
\label{tab:choice_hp}
\end{table*}

All models were trained and evaluated under a unified and controlled experimental pipeline developed for the GSS dataset. To ensure reproducibility, all experiments were conducted with a fixed random seed of \texttt{42}. Default hyperparameters were used for all baselines unless specified otherwise. 

For our proposed CLIP-MHAdapter, the multi-head adapter module was finetuned while keeping the CLIP visual encoder (ViT-B/16) and text encoder (12-layer Transformer) frozen. Images were processed as sequences of $16 \times 16$ patches. We hand-crafted tailored prompts for the dataset based on its attribute set. For example, "a photo taken on {CLASS} is adopted as the prompt template for Platform attribute set". The detailed prompt template design can be seen in the appendix.
To fully leverage multi-head self-attention within the adapter, we employed attribute-specific configurations, with detailed hyperparameters listed in Table~\ref{tab:choice_hp} which were selected according to hyperparameter searching result (see appendix). Training used the AdamW optimiser~\citep{loshchilov2017decoupled} with an initial learning rate of $5 \times 10^{-4}$ and a cosine annealing schedule. A five-epoch linear warm-up was applied, gradually increasing the learning rate from $1 \times 10^{-5}$ to its peak before decaying it following a cosine curve. This scheduling provides a smooth transition from early exploration to more stable convergence. Training was capped at 100 epochs. To mitigate overfitting, early stopping with a patience of 10 validation epochs was employed. If validation accuracy failed to improve within this interval, training was terminated and the best-performing checkpoint retained. 

This approach balances the need to control short-term fluctuations while ensuring efficient convergence. All experiments were conducted on a Linux-based workstation equipped with an NVIDIA RTX 4090 GPU (CUDA 12.8), a 16-vCPU Intel Xeon Gold 6430 processor, and Python 3.10. This hardware configuration provides ample computational capacity for adapter-based finetuning whilst maintaining stable and reproducible training.

\subsection{Evaluation Metrics}\label{sec:evaluation}

To ensure a rigorous and multifaceted assessment for image classification, several metrics were employed: accuracy, macro-averaged F1 score, weighted F1 score, and adjusted balanced accuracy. These capture both global predictive ability and fairness across heterogeneous classes. Table~\ref{tab:symbols} summarises the notation used.

\begin{table}[ht]
\centering
\caption{Symbol glossary for evaluation metric formulas.}
\label{tab:symbols}
\begin{tabular}{ll}
\toprule
\textbf{Symbol} & \textbf{Definition} \\
\midrule
$K$  & Total number of classes \\
$P_i$ & Precision for class $i$ \\
$R_i$ & Recall for class $i$ \\
$n_i$ & Number of true samples in class $i$ \\
$N_s$   & Total number of samples across all classes \\
\bottomrule
\end{tabular}
\end{table}

The most common metric is classification accuracy defined as the proportion of correctly predicted instances over the total. However, accuracy can be misleading under severe class imbalance, as it may overstate performance when one class dominates.

To better account for per-class behaviour, the macro-averaged F1 score is used. As shown in Equation~\ref{eq:macro-f1}, this measure computes the F1 score for each class individually and then averages them with equal weight:
\begin{equation}
\text{Macro-F1} = \frac{1}{K} \sum_{i=1}^{K} \frac{2 \cdot P_i \cdot R_i}{P_i + R_i}
\label{eq:macro-f1}
\end{equation}
This metric ensures that both minority and majority classes contribute equally, thereby reflecting fairness across categories.

A variant of this is the weighted F1 score, defined in Equation~\ref{eq:weighted-f1}. Instead of equal weights, class-specific F1 scores are weighted by the proportion of true samples:
\begin{equation}
\text{Weighted-F1} = \sum_{i=1}^{K} \frac{n_i}{N_s} \cdot \frac{2 \cdot P_i \cdot R_i}{P_i + R_i}
\label{eq:weighted-f1}
\end{equation}
This balances sensitivity to minority classes with representation of dominant ones, offering robustness under skewed class distributions.

Finally, the adjusted balanced accuracy \cite{guyan2015adjustedbalancedaccuracy} corrects for the fact that a random classifier achieves an expected score of $\frac{1}{K}$. As given in Equation~\ref{eq:aba}, the metric rescales the balanced accuracy so that random guessing corresponds to 0 and perfect classification to 1:
\begin{equation}
\text{ABA} = \frac{\frac{1}{K} \sum_{i=1}^{K} \frac{TP_i}{TP_i + FN_i} - \frac{1}{K}}{1 - \frac{1}{K}}
\label{eq:aba}
\end{equation}

Taken together, these metrics provide a comprehensive and robust evaluation framework, particularly suited to the class imbalance and heterogeneous visual conditions present in crowdsourced street-view imagery.

\section{Results and Discussion}
\label{sec:results}

\begin{table*}[htbp]
\centering
\caption{Performance across contextual attributes organised by model paradigm described in \ref{sec:baselines}. \# T. Params denotes the total number of trainable parameters. All metrics are reported as percentages. }
\resizebox{0.95\textwidth}{!}{
\renewcommand{\arraystretch}{1.15}
\begin{tabular}{llllcccc}
\toprule
\textbf{Contextual Attribute} & \textbf{Paradigm} & \textbf{Model} & \textbf{\# T. Params} 
& \textbf{Acc.} & \textbf{Macro F1} & \textbf{Weighted F1} & \textbf{Bal. Acc.} \\
\midrule

\multirow{7}{*}{\textbf{Glare}} 
 & \multirow{2}{*}{Zero-shot Transfer}
 & ZeroR-Trainer & -- & 97.21 & 49.29 & 95.84 & 0.00 \\
 &  & Zero-shot CLIP & -- & 3.03 & 2.96 & 0.62 & 0.24 \\
\cdashline{2-8}
 & Vision Transformer & MaxViT & 30.9M & 94.09 & 63.15 & 95.03 & 39.59 \\
\cdashline{2-8}
 & \multirow{4}{*}{Parameter-Efficient Adaptation}
 & CLIP-Linear Probe & 3K & 95.51 & 53.61 & 95.24 & 6.48 \\
 &  & CoOp & 8K & 96.60 & 57.27 & 95.98 & 10.89 \\
 &  & CLIP-Adapter & 0.52M & 84.16 & 53.65 & 89.16 & 39.26 \\
\cdashline{3-8}
 &  & CLIP-MHAdapter & 1.38M & \textbf{95.32} & \textbf{63.68} & 95.69 & 32.63 \\
\midrule

\multirow{7}{*}{\textbf{Lighting Condition}} 
 & \multirow{2}{*}{Zero-shot Transfer}
 & ZeroR-Trainer & -- & 64.66 & 26.18 & 50.79 & 0.00 \\
 &  & Zero-shot CLIP & -- & 95.88 & 87.65 & 95.45 & 76.54 \\
\cdashline{2-8}
 & Vision Transformer & MaxViT & 30.9M & 96.23 & 90.55 & 96.15 & 84.50 \\
\cdashline{2-8}
 & \multirow{4}{*}{Parameter-Efficient Adaptation}
 & CLIP-Linear Probe & 3K & 89.48 & 69.22 & 88.67 & 55.07 \\
 &  & CoOp & 8K & 94.77 & 81.50 & 93.92 & 68.23 \\
 &  & CLIP-Adapter & 0.52M & 93.57 & 82.91 & 93.51 & 74.96 \\
\cdashline{3-8}
 &  & CLIP-MHAdapter & 1.38M & \textbf{96.46} & 90.29 & \textbf{96.35} & 83.83 \\
\midrule

\multirow{7}{*}{\textbf{Panoramic Status}} 
 & \multirow{2}{*}{Zero-shot Transfer}
 & ZeroR-Trainer & -- & 95.49 & 48.85 & 93.28 & 0.00 \\
 &  & Zero-shot CLIP & -- & 11.92 & 11.85 & 14.18 & 7.76 \\
\cdashline{2-8}
 & Vision Transformer & MaxViT & 30.9M & 99.95 & 99.73 & 99.95 & 99.95 \\
\cdashline{2-8}
 & \multirow{4}{*}{Parameter-Efficient Adaptation}
 & CLIP-Linear Probe & 3K & 87.75 & 67.79 & 90.86 & 87.17 \\
 &  & CoOp & 8K & 98.94 & 94.32 & 98.98 & 95.97 \\
 &  & CLIP-Adapter & 0.52M & 93.69 & 77.60 & 94.87 & 92.42 \\
\cdashline{3-8}
 &  & CLIP-MHAdapter & 1.38M & 99.40 & 96.70 & 99.42 & 98.40 \\
\midrule

\multirow{7}{*}{\textbf{Platform}}
 & \multirow{2}{*}{Zero-shot Transfer}
 & ZeroR-Trainer & -- & 31.69 & 8.02 & 15.25 & 0.00 \\
 &  & Zero-shot CLIP & -- & 60.98 & 43.19 & 60.80 & 45.99 \\
\cdashline{2-8}
 & Vision Transformer & MaxViT & 30.9M & 68.28 & 56.69 & 69.21 & 49.87 \\
\cdashline{2-8}
 & \multirow{4}{*}{Parameter-Efficient Adaptation}
 & CLIP-Linear Probe & 3K & 63.14 & 52.88 & 64.20 & 66.11 \\
 &  & CoOp & 8K & 65.04 & 58.82 & 61.64 & 65.82 \\
 &  & CLIP-Adapter & 0.52M & 68.12 & 57.15 & 69.21 & 71.44 \\
\cdashline{3-8}
 &  & CLIP-MHAdapter & 1.38M & \textbf{69.12} & \textbf{60.79} & 67.27 & 64.93 \\
\midrule

\multirow{7}{*}{\textbf{Quality}}
 & \multirow{2}{*}{Zero-shot Transfer}
 & ZeroR-Trainer & -- & 90.84 & 31.73 & 86.48 & 0.00 \\
 &  & Zero-shot CLIP & -- & 7.40 & 7.32 & 8.07 & 1.43 \\
\cdashline{2-8}
 & Vision Transformer & MaxViT & 30.9M & 79.88 & 40.95 & 83.41 & 27.32 \\
\cdashline{2-8}
 & \multirow{4}{*}{Parameter-Efficient Adaptation}
 & CLIP-Linear Probe & 3K & 86.57 & 53.18 & 87.41 & 33.23 \\
 &  & CoOp & 8K & 92.03 & 42.96 & 89.79 & 11.56 \\
 &  & CLIP-Adapter & 0.52M & 78.69 & 50.80 & 82.99 & 43.80 \\
\cdashline{3-8}
 &  & CLIP-MHAdapter & 1.38M & 89.08 & \textbf{61.46} & 89.62 & 43.78 \\
\midrule

\multirow{7}{*}{\textbf{Reflection}}
 & \multirow{2}{*}{Zero-shot Transfer}
 & ZeroR-Trainer & -- & 72.58 & 42.06 & 61.05 & 0.00 \\
 &  & Zero-shot CLIP & -- & 60.26 & 46.35 & 58.69 & -6.37 \\
\cdashline{2-8}
 & Vision Transformer & MaxViT & 30.9M & 78.72 & 75.67 & 79.56 & 57.61 \\
\cdashline{2-8}
 & \multirow{4}{*}{Parameter-Efficient Adaptation}
 & CLIP-Linear Probe & 3K & 74.94 & 68.19 & 74.81 & 36.02 \\
 &  & CoOp & 8K & 74.66 & 58.75 & 70.32 & 17.10 \\
 &  & CLIP-Adapter & 0.52M & 58.75 & 45.90 & 57.81 & -7.70 \\
\cdashline{3-8}
 &  & CLIP-MHAdapter & 1.38M & 76.69 & 64.93 & 74.10 & 26.97 \\
\midrule

\multirow{7}{*}{\textbf{View Direction}}
 & \multirow{2}{*}{Zero-shot Transfer}
 & ZeroR-Trainer & -- & 88.52 & 46.95 & 83.13 & 0.00 \\
 &  & Zero-shot CLIP & -- & 37.77 & 35.62 & 44.69 & 16.52 \\
\cdashline{2-8}
 & Vision Transformer & MaxViT & 30.9M & 87.38 & 77.99 & 89.06 & 82.35 \\
\cdashline{2-8}
 & \multirow{4}{*}{Parameter-Efficient Adaptation}
 & CLIP-Linear Probe & 3K & 89.51 & 76.96 & 90.06 & 60.65 \\
 &  & CoOp & 8K & 92.89 & 80.87 & 92.55 & 56.56 \\
 &  & CLIP-Adapter & 0.52M & 87.57 & 76.29 & 88.89 & 69.39 \\
\cdashline{3-8}
 &  & CLIP-MHAdapter & 1.38M & \textbf{95.28} & \textbf{87.95} & \textbf{95.19} & 73.19 \\
\midrule

\multirow{7}{*}{\textbf{Weather}}
 & \multirow{2}{*}{Zero-shot Transfer}
 & ZeroR-Trainer & -- & 23.90 & 7.72 & 9.22 & 0.00 \\
 &  & Zero-shot CLIP & -- & 74.43 & 69.33 & 74.13 & 77.95 \\
\cdashline{2-8}
 & Vision Transformer & MaxViT & 30.9M & 75.47 & 59.90 & 74.18 & 51.04 \\
\cdashline{2-8}
 & \multirow{4}{*}{Parameter-Efficient Adaptation}
 & CLIP-Linear Probe & 3K & 57.04 & 59.39 & 56.78 & 56.80 \\
 &  & CoOp & 8K & 84.87 & 85.92 & 84.82 & 82.64 \\
 &  & CLIP-Adapter & 0.52M & 88.01 & 87.69 & 88.08 & 86.72 \\
\cdashline{3-8}
 &  & CLIP-MHAdapter & 1.38M & 81.84 & 85.08 & 82.04 & 83.67 \\
\bottomrule
\end{tabular}
}

\label{tab:results}
\end{table*}

Table \ref{tab:results} summarizes model performance across eight attributes, highlighting that among seven comparable methods CLIP-MHAdapter consistently achieves competitive performance across most contextual attributes.

\subsection{Quantative Analysis}
\label{qual}
CLIP-MHAdapter achieves the best performance under at least one evaluation metric in five out of eight contextual attributes, demonstrating consistently strong effectiveness across diverse conditions. Under the challenging Glare setting, it achieves the highest Macro-F1 (63.68\%) and Weighted-F1 (95.69\%) among parameter-efficient methods, matching the fully fine-tuned MaxViT (30.9M parameters) in balanced recognition. For Lighting Condition, CLIP-MHAdapter attains the best overall accuracy (96.46\%) and Weighted-F1 (96.35\%), slightly surpassing MaxViT and outperforming other Parameter-Efficient Adaptation approaches. On Panoramic Status, its performance achieves the second best across all metrics with 99.40\% accuracy and 96.70\% Macro-F1, approaching the MaxViT’s near-ceiling performance while significantly exceeding Linear Probe and CLIP-Adapter by more than 20–30\% Macro-F1. Similar trends are observed for Platform and Quality, where CLIP-MHAdapter yields the strongest or second-best Macro-F1 scores (60.79\% and 61.46\%, respectively), highlighting improved robustness under class imbalance. For Reflection and View Direction, it consistently improves over other lightweight baselines, achieving 64.93\% and 87.95\% Macro-F1, respectively, and setting the best overall results for view-direction recognition. 
For the weather attribute classification task, CLIP-MHAdapter remains highly competitive, outperforming both zero-shot transfer models and the vision transformer baseline MaxViT, while trailing slightly behind CoOp and CLIP-Adapter. Overall, these results indicate that CLIP-MHAdapter offers an excellent accuracy–efficiency trade-off, delivering performance comparable to fully training on MaxViT with orders-of-magnitude fewer trainable parameters and consistently outperforming existing parameter-efficient adaptation strategies.

\subsection{Qualitative Analysis}
\begin{figure*}[t]
    \centering
    
    \begin{subfigure}{0.8\linewidth}
        \centering
        \includegraphics[width=\linewidth]{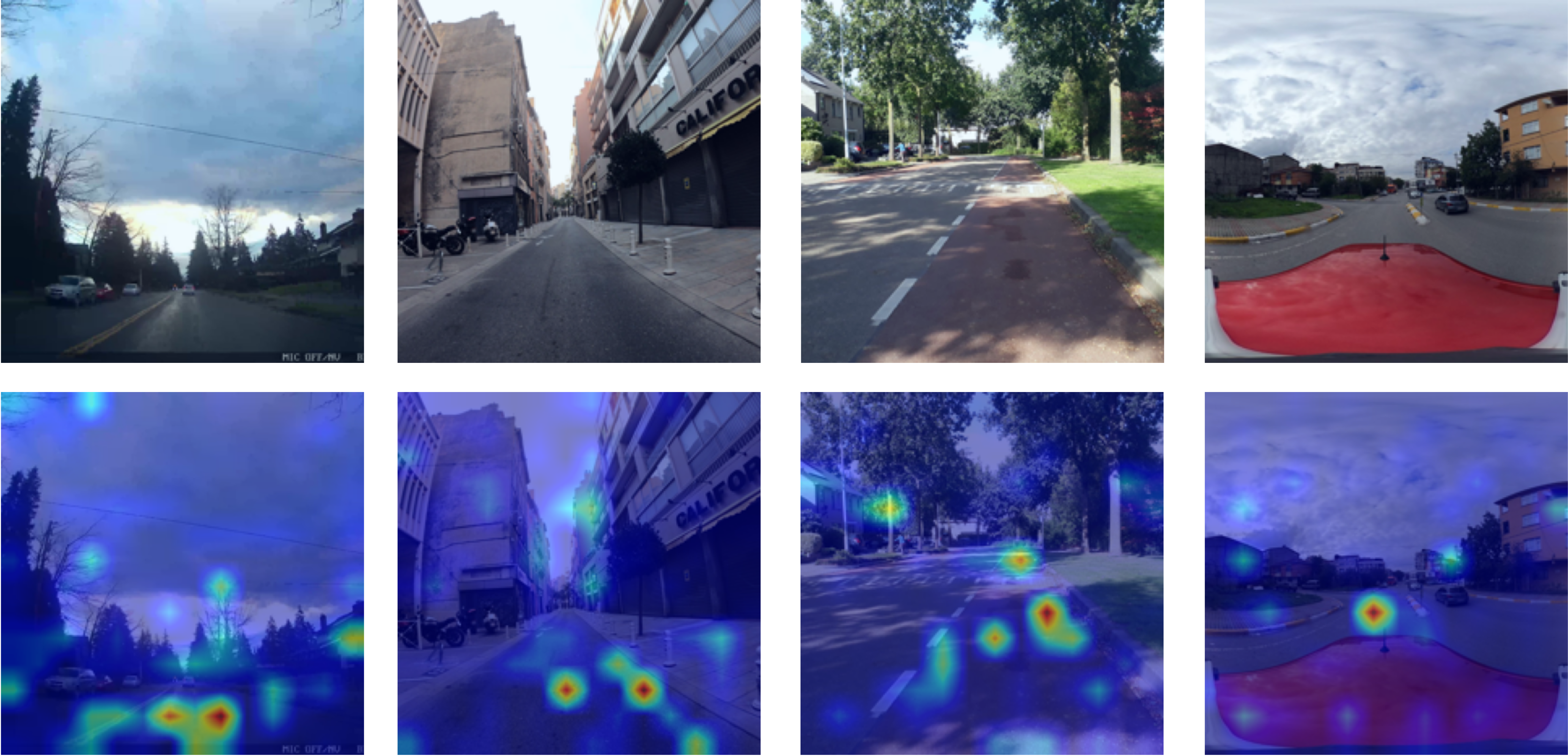}
        \caption{Images with \textbf{Platform} attributes and corresponding attention map.}
    \end{subfigure}

    \vspace{1em} 
    
    \begin{subfigure}{0.8\linewidth}
        \centering
        \includegraphics[width=\linewidth]{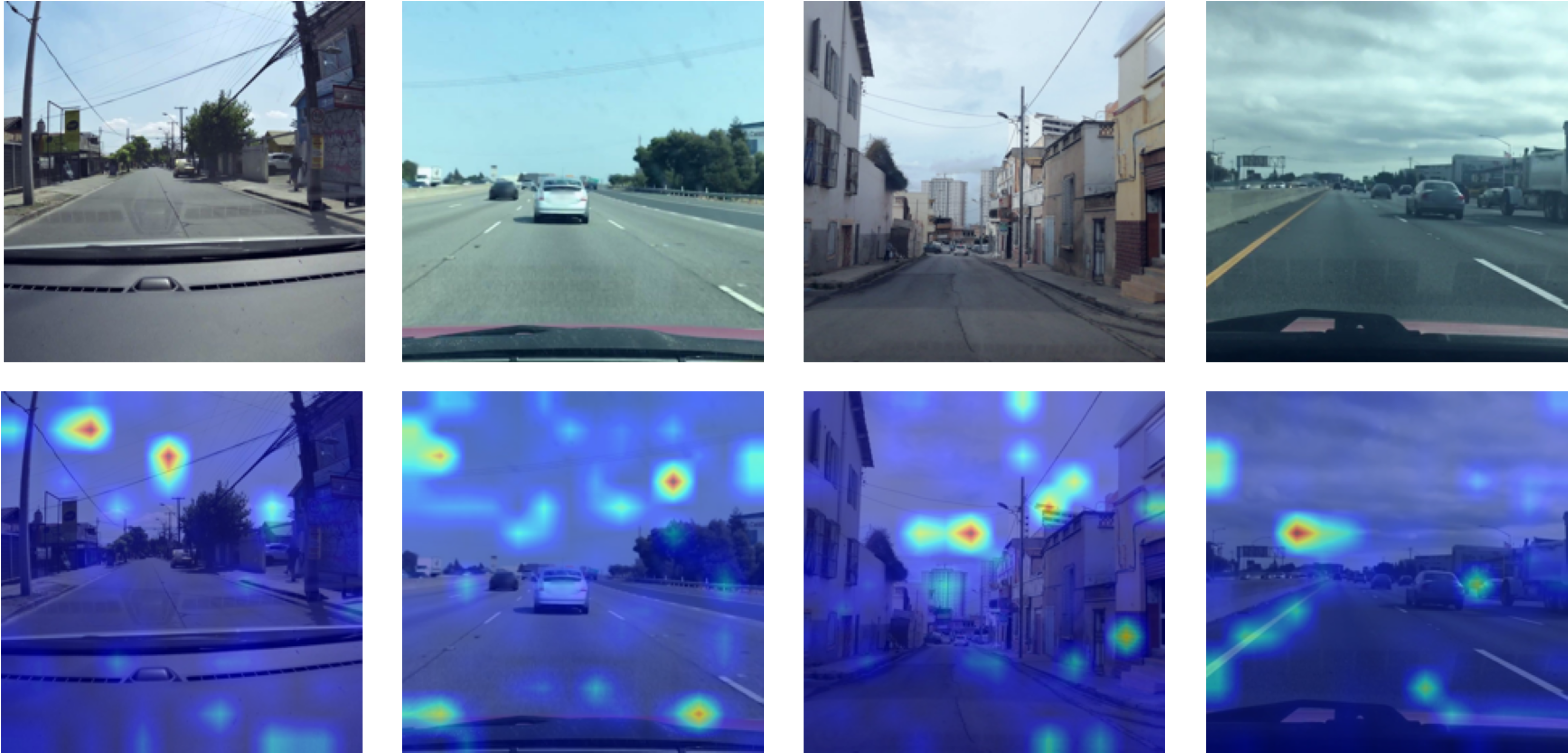}
        \caption{Images with 
        \textbf{Weather} attributes and corresponding attention map.}
    \end{subfigure}

    \caption{Qualitative results of CLIP-MHAdapter. The attention maps from the MHSA layer in the Visual MHAdapter are overlaid on the original input images for visualization.}
    \label{fig:attention-map}
\end{figure*}

The proposed CLIP-MHAdapter achieves superior performance through its multi-head feature adaptation module. In contrast to CLIP-Adapter \cite{gao2024clip}, which employs only an MLP for feature adaptation, CLIP-MHAdapter introduces a multi-head self-attention (MHSA) layer appended after the MLP. This MHSA layer enables the model to selectively attend to informative local spatial features that align with ground-truth attributes during fine-tuning. As illustrated in Figure \ref{fig:attention-map}, the Visual MHAdapter adaptively attends to task-relevant regions in accordance with the target classification attributes. For the Platform attribute classification task, the MHSA layer is trained to emphasise spatial features on the ground, whereas for the Weather classification task, it primarily focuses on spatial patterns in the sky.

\subsection{Efficiency Comparison}

From a comparative standpoint, CLIP-MHAdapter advances the growing line of research on adapter-based multimodal learning. Unlike earlier adapters, it integrates multi-head self-attention at the adaptation stage, improving the capture of inter-token dependencies within visual patch embeddings. This addresses a limitation of lighter adaptation methods, which often struggle to preserve fine-grained spatial relationships without full model fine-tuning.

Relative to alternative approaches, CLIP-MHAdapter achieves a distinctive balance. Zero-shot CLIP, while strong in general image-text alignment, lacked supervision for domain-specific street-view attributes. Prompt-based methods such as CoOp improved adaptability through learnable prompts but remained limited in modelling patch-level interactions. CLIP-MHAdapter, in contrast, delivers competitive performance while being exceptionally lightweight. It requires only 1.38 million trainable parameters which is nearly two orders of magnitude less than MaxViT.
This combination of effectiveness and compactness is a defining characteristic of CLIP-MHAdapter. By maintaining strong accuracy while drastically reducing parameter and resource requirements, it demonstrates that architectural design can achieve a balance between fine-grained discriminative power and efficiency. 

\begin{figure*}[t]
  \centering
  \includegraphics[width=\textwidth]{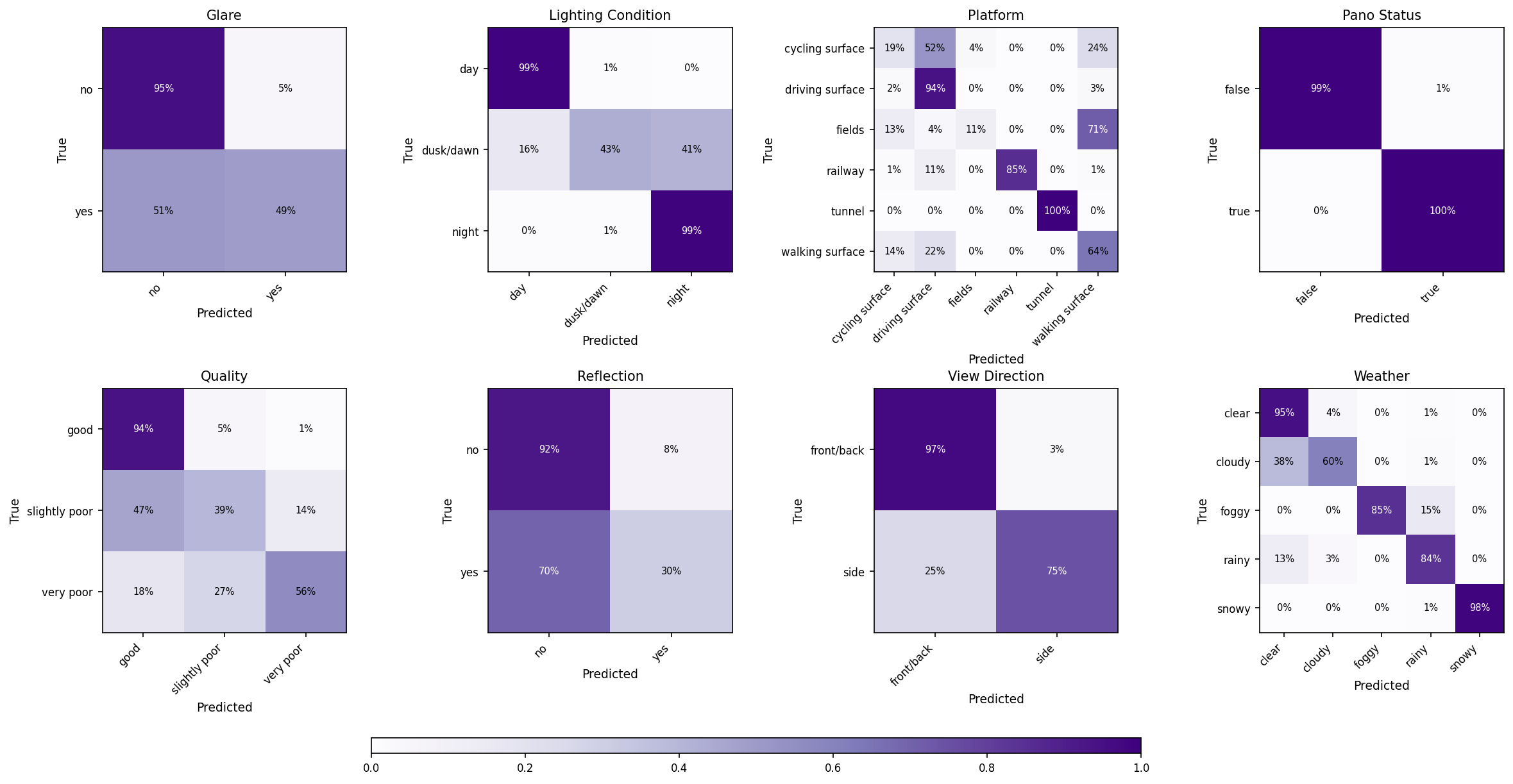}
  \caption{Confusion matrices of CLIP-MHAdapter on the GSS test set across eight attributes classification tasks. Each matrix corresponds to one attribute. }
  \label{fig:confusion}
\end{figure*}

\begin{figure*}[t]
  \centering
  \includegraphics[width=\textwidth]{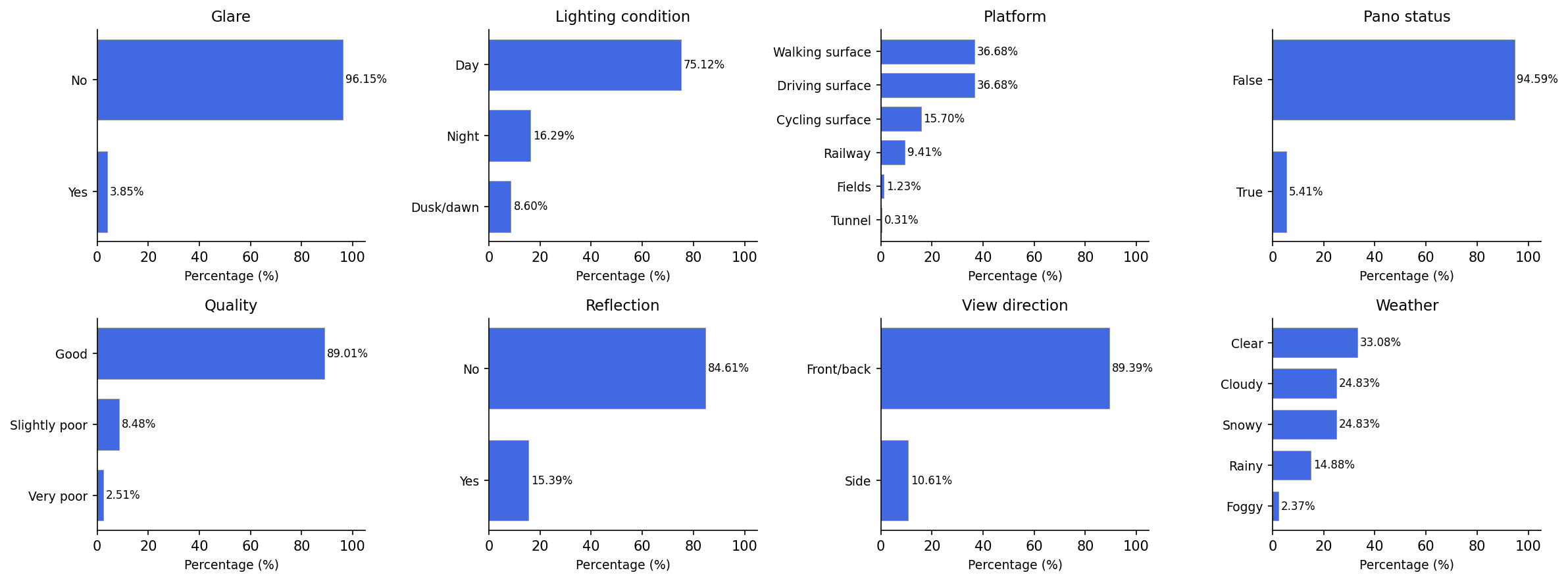}
  \caption{The class distribution for each SVI attribute of labelled GSS dataset \cite{hou2024global}.}
  \label{fig:dist}
\end{figure*}

\subsection{Performance Constraints on GSS Dataset}
Despite its strong overall efficiency–performance trade-off, CLIP-MHAdapter does not uniformly dominate across all contextual attributes. Two dataset-specific constraints in GSS introduce clear limitations that affect reliability and class-wise generalization. First, the GSS dataset exhibits substantial class imbalance across multiple attributes as shown in Figure \ref{fig:dist}, which systematically biases the model toward majority categories and limits its ability to generalize to under-represented cases. For example, the glare attribute is overwhelmingly dominated by non-glare samples, accounting for approximately 96 percent of the data, while similar skewed distributions are observed for panoramic status, where the vast majority of samples correspond to the negative class, image quality, where most images are labelled as good quality, reflection, where no reflection takes up over 84 percent of data, and view direction, where front- or back-facing views constitute nearly nine-tenths of the dataset. Such imbalance might cause model overfitting to dominant classes during optimisation, resulting in noticeably reduced recall for minority categories for the classification task of the aforementioned attributes.

Particularly, for Weather, although competitive, CLIP-MHAdapter underperforms CLIP-Adapter and CoOp (as discussed in \ref{qual}), misclassifying 38 percent 'cloudy' images as 'clear' as shown in Figure \ref{fig:confusion}. For Reflection, CLIP-MHAdapter achieves moderate improvements over zero-shot and linear probing baselines but remains noticeably below the fully fine-tuned MaxViT, misclassifying 70 percent images containing reflection as non-reflective.
Arguably, the relatively low inter-annotator agreement across certain attributes in the original labelling process of the GSS dataset may introduce label noise, thereby further contributing to the observed performance degradation.
According to \cite{hou2024global}, the lowest agreement of 66.5\% in weather labelling among all attributes often stemmed from the difficulty of determining whether an image should be considered cloudy or clear given the level of cloud cover. The second lowest agreement among eight attributes is Reflection. This implies that the model’s performance on the weather attribute may not accurately reflect its true capability, largely due to ambiguous ground-truth labels.



\section{Conclusion}
\label{sec:conclusion}

This work investigated the challenge of fine-grained street-view image attribute classification, a task crucial for autonomous driving, urban analytics, and large-scale mapping. To address this, CLIP-MHAdapter, a lightweight residual-style adapter enhanced with multi-head self-attention, was proposed to leverage CLIP’s pretrained representations while avoiding the computational cost of full fine-tuning.

Empirical evaluation showed that CLIP-MHAdapter consistently outperformed zero-shot CLIP, CoOp, and the baseline CLIP-Adapter across multiple metrics. It also delivered competitive results compared to state-of-the-art vision transformers such as MaxViT, while requiring significantly fewer trainable parameters and computational resources. These findings demonstrate that adapter-based strategies, when augmented with self-attention, can effectively combine efficiency with fine-grained discriminative power. More broadly, this research highlights the potential of lightweight adapter-based approaches as a scalable pathway for transferring vision-language foundation models like CLIP to specialised, real-world domains.

{
    \small
    \bibliographystyle{ieeenat_fullname}
    \bibliography{main}
}


\end{document}